\documentclass{llncs}

\usepackage[T1]{fontenc} 
\usepackage{tgtermes}
\usepackage{amssymb}
\usepackage{amsmath}
\usepackage{graphicx}
\usepackage{esint}
\usepackage{enumitem}
\usepackage{algorithm}
\usepackage{fancyvrb}
\usepackage{hyperref}
\usepackage{marvosym}
\usepackage{mathtools}
\usepackage{float}
\usepackage{xspace} 
\usepackage{color}
\usepackage{comment}
\usepackage{listings}
\usepackage{siunitx}

\usepackage[noend]{algpseudocode}
\usepackage[section]{placeins}
\usepackage{hyperref}

\algrenewcommand\algorithmicindent{1.0em}

\newcommand\rcode[1]{\textcolor[rgb]{1,0,0}{\textbf{#1}}}

\newcommand\bcode[1]{\textcolor[rgb]{0,0,1}{\textbf{#1}}}

\begin{document}
\title{Accelerated Labeling of Discrete Abstractions for Planning Under LTL Specifications}

\author{Brian Paden, Peng Liu, and Schuyler Cullen}
\institute{Advanced Technology Group @ Samsung Smart Machines \\ 
\email{brian.paden@samsung.com}, $\quad$ \email{peng.liu@samsung.com}, \quad \email{s.cullen@samsung.com}}
\maketitle
\begin{abstract}
Linear temporal logic and automaton-based runtime verification provide a powerful framework for designing task and motion planning algorithms for autonomous agents.
The drawback to this approach is the computational cost of operating on high resolution discrete abstractions of continuous dynamical systems.
In particular, the computational bottleneck that arises is converting perceived environment variables into a labeling function on the states of a kripke structure or similarly the transitions of a labeled transition system.
This paper presents the design and empirical evaluation of an approach to constructing the labeling function that exposes a large degree of parallelism in the operation as well as efficient memory access patterns.
The approach is implemented on a commodity GPU and empirical results demonstrate the efficacy of the labeling technique for real-time planning and decision-making.

\end{abstract}

\section{Introduction}
	In the context of autonomous systems and robotics, linear temporal logic (LTL) provides an expressive language for defining desired properties of an autonomous agent that extends the typical point-to-point motion planning problem which has been studied extensively in the robotics literature~\cite{lavalle2006planning}. 
	LTL extends predicate logic with operators that allow constraints to be placed on the ordering of events. 
	The techniques that have been developed for planning motions satisfying specifications given as LTL formulae are well suited to systems requiring guaranteed satisfaction of safety requirements and traceability of failures.

	Most approaches share two key elements: (i) a discrete abstraction of the system is constructed along with a labeling of the discrete system with properties relevant to the specification, (ii)  the specification is translated into a finite automaton which accepts runs of the system satisfying the specification.
	However, practical challenges lead to many variations on this approach. 
	Discrete abstractions can be constructed by sampling a finite number of discrete motions~\cite{karaman2011sampling,vasile2013sampling} in the state space or by a finite partition of the state space~\cite{kloetzer2008fully,kress2009temporal}.
	The discrete abstraction was formulated as a Markov Decision process to handle uncertainty in the system in addition to a performance objective in~\cite{ding2014optimal}. Resolving conflicting specifications was treated in~\cite{tuumova2013minimum}, and generating trajectories satisfying LTL formulae using Monte Carlo methods was recently investigated in~\cite{paxton2017combining}. 

	While there are numerous variations on the basic approach available, there are many remaining challenges related to bringing the theory to practice.
	A recent literature review~\cite{plaku2016motion} discusses some of the open problems in greater detail,
	One of the principal challenges with planning to meet an LTL specification is the labeling of the discrete system with features extracted from sensor data by the perception system.
	These geometric computations dominate the computational requirements of the planning process.
	This paper investigates practical aspects of constructing the labeling function in real-time by leveraging precomputed data and a highly parallel implementation.
	A fixed discrete abstraction is computed offline and used in multiple receding horizon planning queries at run-time by exploiting equivariace in the differential constraints present in models for mobile agents. Equivariance in the dynamic model was similarly exploited in~\cite{frazzoli2005maneuver}.
	Data from the perception system at a given instant can then be mapped, in parallel, to each of the precomputed trajectories making up the discrete abstraction.
	
	Constructing discrete abstractions for differentially constrained systems is reviewed in Section \ref{sec:discrete_abstraction}. 
	Section \ref{sec:properties} discusses labeling of discrete abstractions with properties relevant to the task specification and construction of monitors to detect violation of safety properties expressed in LTL.
	The principal contribution of the paper, presented in Section \ref{sec:acceleration}, describes a reduction of the labeling operation to a binary matrix multiplication with efficient processor memory access patterns and is readily accelerated by parallel computation on a graphics processing unit (GPU) or application specific integrated circuit. 
	Lastly, the labeling technique is tested on a probabilistic roadmap~\cite{karaman2011sampling} designed for autonomous driving using on a commodity GPU.

\section{\label{sec:discrete_abstraction}Discrete Approximations of Robot Mobility}
The mobility of an autonomous agent is initially modeled by a controlled dynamical system with \emph{state} $x(t)\in \mathbb{R}^n$ and \emph{control action} $u(t)\in \mathbb{R}^m$ at time $t\in \mathbb{R}$.
The dynamical system, derived from first principles, relates control actions to the resulting trajectory
\begin{equation}\label{eq:dynamics}
\frac{d}{dt}x(t)=f(x(t),u(t)).
\end{equation}
An initial condition $x(t_0)=x_0$ is given, and the control actions $u:[t_0,t_f]\rightarrow \mathbb{R}^m$ must be selected so that the unique trajectory through $x_0$ resulting from taking actions $u$ meets a planning specification and minimizes a cost function $J(x,u)$ of the form
\begin{equation}
C(x,u) = \int_{t_0}^{t_f} g(x(t),u(t))\, dt.
\end{equation}
%
%

In the interest of computing a motion meeting the specification in real-time, many approaches to motion planning approximate the set of  
trajectories satisfying \eqref{eq:dynamics} as a directed graph or \emph{transition system} $(V,E)$ where the $V$ is a finite subset of the state space containing $x_0$, and $E\subset V\times V$. 
For each transition $(v_a,v_b)\in E$, we  associate a trajectory and control signal $(x,u)$ with finite duration $[t_a,t_b)$ such that $x(t_a)=v_a$ and $\lim_{t\rightarrow t_b} x(t) = v_b$. The trajectory and control signal associated to a transition $e\in E$ will be denoted $X(e)$ and $U(e)$ respectively.
Similarly, the net cost of an edge is given by $C(X(e),U(e))$, which will be abbreviated with some abuse of notation by $C(e)$.

A \emph{finite trace} $(v_0,v_1,...,v_n)$ of the transition system is a finite sequence of states in $V$ such that each sequential state is a transition $(v_i,v_{i+1})\in E$.   
Due to the time invariance of \eqref{eq:dynamics}, a feasible trajectory and control signal can be recovered from each finite trace by concatenating the trajectories and controls associated to each transition $X((v_0,v_1))X((v_1,v_2))...X((v_{n-1},v_n))$, and $U((v_0,v_1))U((v_1,v_2))...U((v_{n-1},v_n))$.

\section{Properties of Trajectories}\label{sec:properties}


A finite set of atomic propositions $\Pi$, rich enough to express the task planning specification in the syntax of LTL, are given. 
For example, in the context of advanced driver assistance systems, $\Pi$ might include $\texttt{DrivableSurface}$, $\texttt{LegalSurface}$, $\texttt{NominalSurface}$, etc.
In a given scenario, an interpretation of true or false ($\top$ or  $\bot$) is assigned to each proposition in $\Pi$ at each state.
A \emph{state labeling function} provides the map from each state to the interpretation of each proposition $\mathcal{L}:\mathbb{R}^n\rightarrow \{\top,\bot\}^{|\Pi|}$ (there is a natural bijection between $\{\top,\bot\}^{|\Pi|}$ and the powerset $2^\Pi$ and it is customary to use the powerset representation).
%

The state labeling function can be applied point-wise to a trajectory $x$ to construct a \emph{state labeling function} $L:E\rightarrow 2^\Pi$ defined as follows
\begin{equation}
L(e) \coloneqq \bigcup_{t\in[t_a,t_b]} \mathcal{L}([X(e)](t)). 
\end{equation}
Intuitively, this construction labels a transition $e$ with each proposition encountered by the associated state trajectory $X(e)$.
Sequences of transitions of the transition system form strings over $2^{\Pi}$ which can be scrutinized for satisfaction or violation  of the task specification.
%
%
%

\subsection{Linear Temporal Logic as a Specification Language}
Linear temporal logic (LTL) has become one of the predominant means of specifying desired properties of motions for autonomous agents.
It consists of the usual logical operators $\neg$ (not), $\wedge$ (and), $\vee$ (or), together with the temporal operators $\mathcal{U}$ (until) and $\bigcirc$ (next).
The set of LTL formulae are defined recursively as follows:
\begin{enumerate}
	\item Each subset of $\Pi$ is a formula.
	\item If $\phi$ is a formula, then $\neg\phi$ is a formula.
	\item If $\phi_1$ and $\phi_2$ are formulae, then $\phi_1 \vee \phi_2$ is a formula.
	\item If $\phi_1$ and $\phi_2$ are formulae, then $\phi_1 \mathcal{U} \phi_2$ is a formula.
	\item If $\phi$ is a formula, then $\bigcirc \phi$ is a formula.
\end{enumerate}

Let $w=w_0w_1w_2...$ be an infinite sequence of elements from $2^{\Pi}$. 
Such a sequence is called an $\omega$ word over $2^{\Pi}$. 
In the context of motion and task planning, each $w_i$ is a subset of $\Pi$ representing the atomic propositions which are true at time $i$.
The semantics of LTL define which words $w$ satisfy a LTL formula $\phi$, in which case we write the relation $w \models \phi$. 
A pair $(w,\phi)$ in the complement of the satisfaction relation is denoted $w\not\models\phi$.
The satisfaction relation for LTL is defined recursively as follows:
\begin{enumerate}
	\item For $p\subset \Pi$,  $w\models p$ if $p \in w_0$.
	\item $w\models \neg\phi$ if $w \not\models \phi$.
	\item $w\models \phi_1 \vee \phi_2$ if $w\models\phi_1$ or $w\models\phi_2$.
	\item $w\models \bigcirc\phi$ if $w_1w_2... \models \phi$.
	\item $w\models \phi_1 \mathcal{U} \phi_2$ if there exists $i$ such that $w_iw_{i+1}...\models\phi_2$ and for each $j<i$, $w_jw_{j+1}...\models\phi_1$.
\end{enumerate}
Useful constructs derived from these operators are the following:
\begin{enumerate}
	\item $\phi_1 \wedge \phi_2 \coloneqq \neg(\neg \phi_1 \vee \neg \phi_2)$
	\item $\phi_1 \Rightarrow \phi_2 \coloneqq \neg \phi_1 \vee \phi_2$
	\item $\phi_1 \Leftrightarrow \phi_2 \coloneqq (\phi_1 \Rightarrow \phi_2) \wedge (\phi_2 \Rightarrow \phi_1$)
	\item $\top \coloneqq \phi \wedge \neg \phi$
	\item $\bot \coloneqq \neg \top$
	\item $\lozenge \phi \coloneqq \top \mathcal{U} \phi$
	\item $\square \phi \coloneqq \neg \lozenge\neg \phi$
\end{enumerate}

The set of $\omega$ words that satisfy a formula $\phi$ is an $\omega$-regular language.
When monitoring a system in real time, only finite prefixes of infinite execution traces are observable from the finite operating time of the system so it cannot always be determined from a finite prefix of an $\omega$-word if it will ultimately satisfy or violate a particular LTL formula.
For example, satisfaction of a persistent surveillance specification, which can be expressed as $\phi = \square \lozenge X$ (always eventually visit X), cannot be verified by observing a finite prefix of an infinite trace since a future visit to $X$ must take place at a later time than what can be observed.
A finite prefix is a \emph{bad prefix} if all $\omega$-words beginning with that prefix fail to satisfy a formula. 
The notion of \emph{good prefix} is defined analogously.
%
%

\subsection{B\"uchi Automata}

A \emph{B\"uchi Automaton} $\mathcal{A}=(Q,\delta,q_0,F)$ consists of a set of states $Q$, a transition relation $\delta \subset Q\times 2^\Pi \times Q$, an initial state $q_0$ and a set of accepting states $F\subset Q$. 
An $\omega$-regular word $w=w_0w_1w_2...$ is \emph{accepted} by a B\"uchi Automaton if there exists a sequence of states $q_0q_1q_2...$ in $Q$ such that $(q_{i},w_i,q_{i+1})\in \delta$ and if there exists a state $q$ in $F$ appearing infinitely often in the sequence of states $q_0q_1q_2...$ associated with $w$.
The purpose of runtime monitors is to identify good and/or bad prefixes from finite executions of a system.

Like the $\omega$-regular words satisfying an LTL formula, the $\omega$-regular words accepted by a B\"uchi Automaton is an $\omega$-regular language.
This language will be nonempty if there are strongly connected components of the automaton reachable from the initial state.
A variety of algorithms and implementations are available (e.g. ~\cite{gastin2001fast}) for constructing a B\"uchi Automaton accepting exactly the $\omega$-regular words satisfying an LTL formula.

A subset of the automaton's states $Q$ of interest are those states which can reach a strongly connected component.
Let $\hat{F}$ denote this subset, and define the nondeterministic finite automaton  $\hat{\mathcal{A}}=(Q,\delta,q_0,\hat{F})$. 
Among the results in~\cite{bauer2011runtime} was the observation that the regular language (in contrast to $\omega$-regular) rejected by this new automaton is exactly the set of bad prefixes violating the safety specification.
This provides a means of detecting violations of LTL safety specifications from finite strings.

Restricted to the discrete abstraction $(V,E)$ with labeling function $L:E\rightarrow 2^{\Pi}$ and cost function $C:E\rightarrow \mathbb{R}$ discussed in Section \ref{sec:discrete_abstraction}, the set of trajectories with finite duration which do not violate a LTL formula $\phi$ are given by paths $v_0v_1v_2...v_n$ on $(V,E)$ such that there exists a sequence of states $q_0q_1...q_n$ on $\hat{\mathcal{A}}$ with $(q_i,L(v_i,v_{i+1}),q_{i+1})\in\delta$. 
Note that these are simply the paths in the product graph with vertices $V \times Q$, and edges $((v_i,q_i),(v_j,q_j))$ in the edge set if $(v_i,v_j)\in E$ and $((q_i,\mathcal{L}(v_i,v_{j}),q_{j}))\in \delta$.
The weighting function on edges of $(V,E)$ can be extended to edges of the product graph as $\tilde{C}((v_i,q_i),(v_j,q_j))\coloneqq C((v_i,v_j))$.
A minimum cost path reaching a terminal state in $V$ without violating the LTL formula $\phi$ can then be computed by solving the shortest path problem on the product graph.

\begin{figure}
	\centering
	\includegraphics[width=1.0\columnwidth]{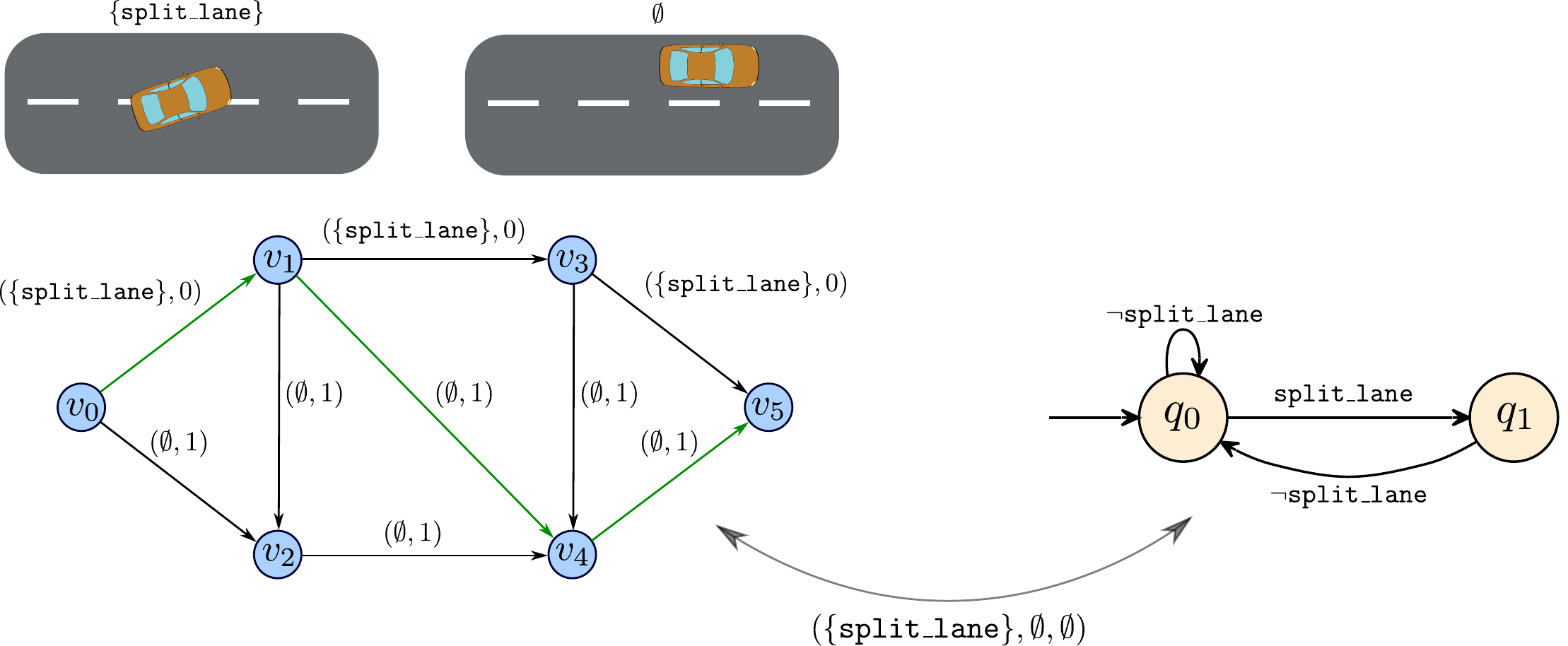}
	\caption{(left) A graphical representation of a labeled transition system with 6 states and 9 transitions. (right) A runtime monitor for the specification in \eqref{eq:split_spec}. Transitions are labeled by elements from the powerset of the singleton $ \{ \texttt{split\_lane} \}$ and a cost of either $0$ or $1$. A minimum cost path not violating \eqref{eq:split_spec} and connecting $v_0$ to $v_5$ on the transition system is given by the shortest path $(v_0,q_0)(v_1,q_1)(v_4,q_0)(v_5,q_0)$ on the product graph. In contrast, a shortest path $v_0v_1v_3v_5$ on the transition system has lower cost but violates the specification. }\label{fig:product_graph}
\end{figure}

\paragraph{Example}
As an example in the context of autonomous freeway driving, the requirement that the vehicle should not split driving lanes for two or more consecutive transitions in the state abstraction can be expressed as 
\begin{equation} \label{eq:split_spec}
\varphi = \square (\texttt{split\_lane}  \Rightarrow \bigcirc \lnot \texttt{split\_lane} ).
\end{equation}
Figure \ref{fig:product_graph} illustrates how this specification can be monitored on traces of a discrete abstraction of the autonomous vehicle's model of mobility. 

\section{\label{sec:acceleration}Real-time Labeling of Transition Systems}
The above discussion outlines a mathematical framework for approaching motion planning problems subject to specifications given as LTL formulae.
In practice, the computational bottleneck encountered in implementing this approach is constructing the labeling function $\mathcal{L}$ determined by the scene perceived by the perception system.
The subset of a state space where an atomic proposition is interpreted as $\top$ is typically defined by states where the associated physical volume occupied by the agent intersects some volume associated with that  proposition.
The physical subset of space\footnote{This is generally a three dimensional space, but a two dimensional space may be sufficient for ground robots. Alternatively, time could be included in a spatio-temporal workspace leading to potentially four dimensional workspaces.} occupied by the autonomous agent is distinct from, and may have differing dimension than the state space.
This is referred to as the \emph{workspace}, and the mapping from a state of the agent to the subset occupied by the agent in the workspace will be denoted by 
\begin{equation}\label{eq:workspace_proj}
F:X \rightarrow 2^W.
\end{equation}
\begin{figure}
\centering
	\includegraphics[width=0.7\columnwidth]{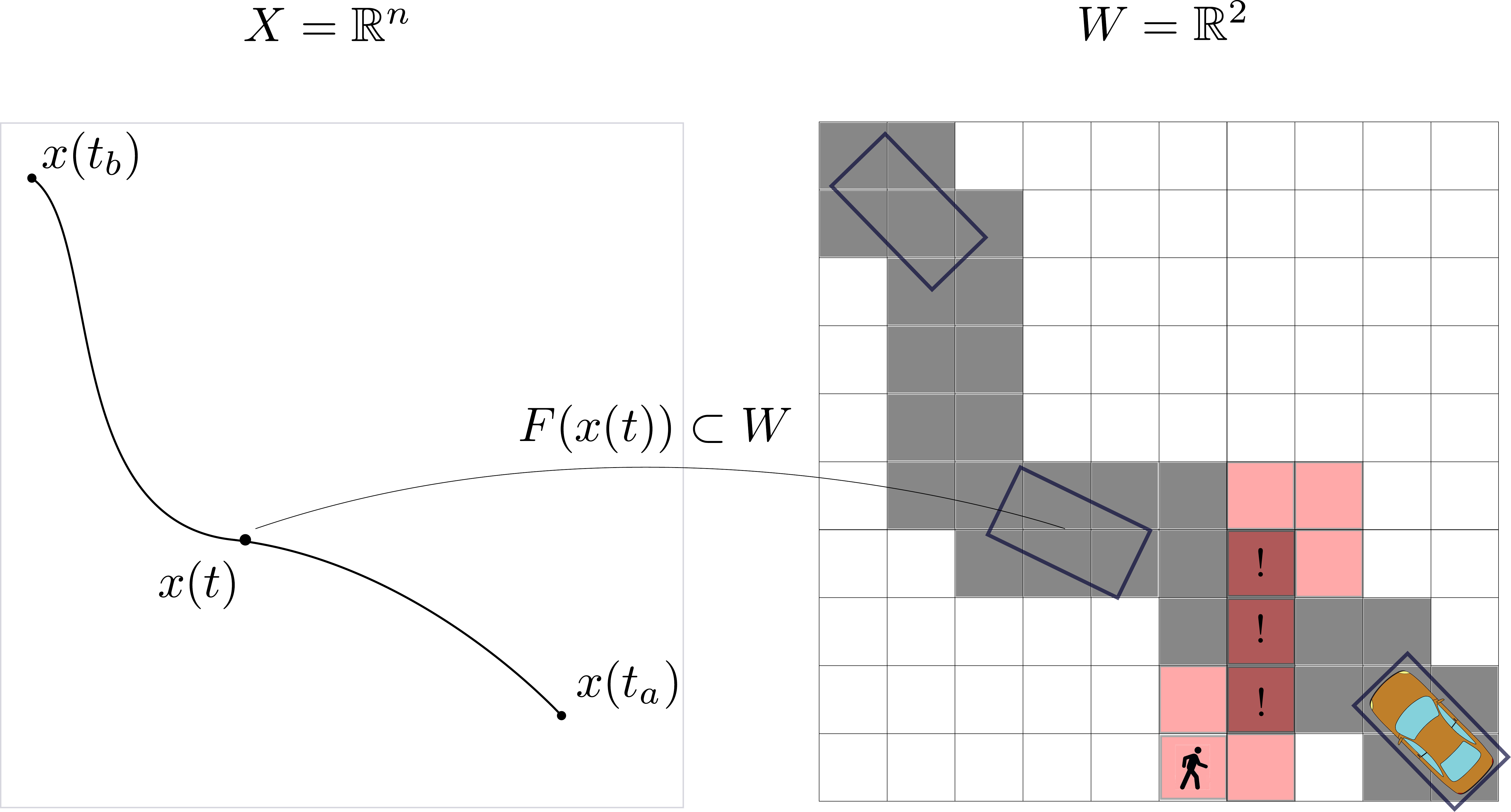}
	\caption{An illustration of a trajectory $x$ within the state space $X$ associated with edge $e$. Each point along the trajectory is mapped by $F$ to a subset of the workspace $W$. In this illustration $F(x(t))$ is the rectangular footprint in a top down view of the car. An atomic proposition $\mathtt{pedestrian}$ associated to the red region $S_{\mathtt{pedestrian}}$ intersects $F(x(t))$ for some values of $t$. Thus, $\mathtt{pedestrian}\in\mathcal{L}(e)$.  }\label{fig:state_to_workspace}
\end{figure}

In this context, an atomic proposition $\pi$ associated to a subset $S_{\pi}$ of the workspace belongs to the label of an edge $e$ if 
\begin{equation}
	\left(\bigcup_{t\in[t_a,t_b]} F([X(e)](t)) \right)\cap S_{\pi} \neq \emptyset. 
	\end{equation}
	This allows for a succinct definition of the labeling function $\mathcal{L}:E\rightarrow 2^{\Pi}$ as follows:
	\begin{equation}\label{eq:labeling_op}
	\pi \in \mathcal{L}(e) \Leftrightarrow \left(\bigcup_{t\in[t_a,t_b]} F([X(e)](t)) \right)\cap S_{\pi} \neq \emptyset.
	\end{equation}
	Figure \ref{fig:state_to_workspace} illustrates the definition of the labeling function.
	The focus of the remainder of the paper is a discussion of how to approximately evaluate \eqref{eq:labeling_op} in real-time.

	\subsection{Partitioning and Indexing the Workspace}\label{sec:partition}
	Subsets associated to particular predicates in the workspace must be approximated by some finite representation. 
	%
	The workspace geometry is restricted to rectangular regions in $[l_1,u_1]\times ... \times [l_k,u_k]\subset \mathbb{R}^k$.
	This region is partitioned into an occupancy grid of hyper-rectangular regions and these cells are indexed with integer values using a $k$-dimensional z-order curve.
	Algorithm \ref{alg:zcurve} illustrates how an index is computed for a particular point in the workspace by descending a space partitioning binary tree to determine the index of a point. 
	Each level of the binary tree represents a contribution of a power of two to the index. 
	If the point is on the high side of a space partitioning hyperplane at depth $i$, then $2^i$ is added to the index.  
	Algorithm \ref{alg:zcurve} indexes points in $\mathbb{R}^k$ along the z-order curve with. 
	With finite precision, the same indexing can be accomplished in finite time by interleaving the bits of the individual coordinates of the point.
	
	\begin{algorithm} 
		\begin{algorithmic}[1]
			\State $z_i \gets (x_i-l_i)/(u_i-l_i),\qquad \forall i\in\{0,...,k-1\}$ \Comment Map workspace to $[0,1]^k$
			\State $j\gets 0$ \Comment Axis used to define splitting plane of workspace
			\State $n \gets 0$ \Comment The initial value for the index
			\State $p\gets (0.5,0.5,...,0.5)$ \Comment Start pivot in center of cube
			\For {$i=1,\dots, d$}\Comment Descend the binary tree to depth $d$ 
			\If $x_j<p_j$\Comment Check if query is on high side of splitting plane
			\State $p_j \gets p_j + 0.5^{\lfloor (i+2k)/k \rfloor}$ \Comment Step in direction of query along current axis 
			\State $n = n + 2^{d-i}$ \Comment Increase index with order of magnitude determined by depth
			\Else 
			\State $p_j \gets p_j - 0.5^{\lfloor (i+2k)/k \rfloor}$ \Comment Step in direction of query along current axis 
			\EndIf
			\State $j \gets j+1$ \Comment Increment splitting plane axis
			\EndFor
			\State \Return $n$
		\end{algorithmic} \caption{\label{alg:zcurve} Input: $x\in \mathbb{R}^k$, and $[l_1,u_1]\times ... \times [l_k,u_k]\subset \mathbb{R}^k$. Output: $n\in \mathbb{N}$ } 
	\end{algorithm}
	
\begin{figure}
		\includegraphics[width=0.6\columnwidth]{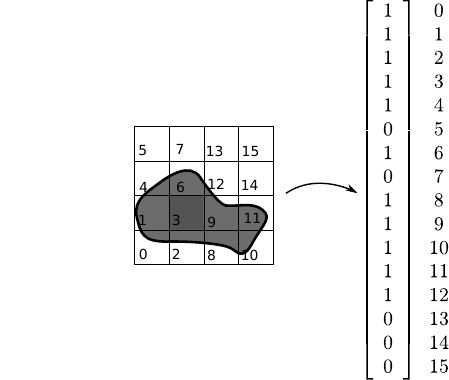}
		\caption{The grey region illustrates a subset of a 2-dimensional workspace. The subset is approximated by the rectangular regions it intersects and encoded as a binary array with a 1 entry at the array index associated to the indices of the intersected regions.}\label{fig:indexing}
	\end{figure}

	By partitioning the workspace and indexing the hyper-rectangular cells, each possible subset of the workspace can be approximated and encoded by a binary vector in $v\in\{0,1\}^{2^d}$ with $v_i=1$ if the subset intersects the region indexed with the value $i$ and $0$ otherwise.   
	This is illustrated in Figure \ref{fig:indexing}. 
	
	\subsection{Labeling the Transition System via Boolean Matrix Multiplication}
	In reference to \eqref{eq:labeling_op}, determining if $\pi \in \mathcal{L}(e)$ requires intersecting two subsets of the workspace. 
	If the subset associated with the trajectory is approximated by the binary vector $T_e$, and the subset associated with the proposition $\pi$ is approximated by a binary vector $\hat{S}_{\pi}$, whether or not their intersection is empty can be computed by evaluating 
	\begin{equation}
	\bigvee_{i} T_e(i) \land \hat{S}_{\pi}(i).
	\end{equation}
	In connection to the labeling function,
	\begin{equation}\label{eq:bool_dot}
	\pi \in \mathcal{L}(e) \Leftrightarrow  \bigvee_{i} T_e(i) \land \hat{S}_{\pi}(i)=\top.
	\end{equation}
	
	Let the matrix $M$ be a binary matrix who's rows are the binary vectors associated to each trajectory in the transition system approximating the autonomous agent's mobility.
	Similarly, let $P$ be a binary matrix who's columns correspond to the binary vectors associated with each atomic proposition.
	Then computation of \eqref{eq:labeling_op} for each proposition and trajectory can be distilled into a binary matrix multiplication,
	\begin{equation}\label{eq:approx}
	L(i,j) = \bigvee_k M(i,k)\land P(k,j),
	\end{equation} 
	where $L(i,j)=\top$ indicates that $\pi_j\in L(e_i)$.

	\subsection{Accelerated Binary Matrix Multiplication}
	The following implementation discussion makes use of several practical observations and assumptions: 
	\begin{enumerate}
		\item[A1] The required number of trajectories and number of rectangular regions partitioning the workspace is large relative to the number of atomic propositions.
		\item[A2] As a result of having a large number trajectories, the duration of each trajectory is small leading to a small fraction of the workspace swept out by $F[(X(e_i)](t)$. In contrast, the volume of the workspace associated to a particular predicate could be large (e.g. the drivable surface in a scene).
		\item[A3] The volume swept out in the workspace by $F[(X(e_i)](t)$ will be a somewhat spatially coherent region which maps to a small number of clustered $\top$ entries (via the z-order curve) in the associated binary vector. In contrast, the volume occupied by an atomic proposition may be a large fraction of the workspace.
	\end{enumerate}
	
	Since each trajectory sweeps out a small fraction of the workspace, the matrix M in \eqref{eq:approx} will be sparse in the sense that it will have few $\top$ entries. 
	This suggests that the matrix $M$ be stored in \emph{compressed sparse row} (CSR) format.
	The location of $\top$ entries of $M$ are represented by two arrays $r$ and $c$. 
	The $i^{th}$ entry of $c$ is the total number of $\top$ entries in rows $0$ up to, but excluding, $i$. The entries of $r$ contain the column index of each $\top$ entry of $M$ read from left to right and top to bottom.
	This is illustrated with the following example matrix and its CSR format: 
	\begin{equation}\label{eq:csr_demo}
	\begin{gathered}
	\begin{array}{c}
	0\\
	1\\
	2\\
	3\\
	4
	\end{array}\overset{\arraycolsep=2.9pt\begin{array}{ccccc}
		0 & 1 & 2 & 3 & 4\end{array}}{\left(\begin{array}{ccccc}
		\bot & \bot & \bot & \bot & {\boldsymbol\top}\\
		\bot & \boldsymbol{\top} & \boldsymbol{\top} & \bot & \bot\\
		\boldsymbol{\top} & \bot & \bot & \bot & \bot\\
		\bot & \bot & \boldsymbol{\top} & \boldsymbol{\top} & \bot\\
		\bot & \bot & \bot & \boldsymbol{\top} & \bot
		\end{array}\right)},\\ \\
	r=(0,1,3,4,6,7),\quad
	c=(4,1,2,0,2,3,3).
	\end{gathered}
	\end{equation} 
	The proposition matrix $P$ is much smaller than the trajectory matrix $M$ and is not observed to be particularly sparse. 
	Therefore, it is left in the dense 2-dimensional array format and the appropriate sparse-dense matrix multiplication algorithm is described in Algorithm \ref{alg:mat_mul}
	\begin{algorithm} 
		\begin{algorithmic}[1] 
			\For {$i=0\,\,to\,\,\#cols(P)$}\Comment Loop over each column of P 
			\For {$j\in r$}\Comment Traverse nonzero entries for each row of $M$
			\For {$k = r(j) \,\, to \,\,r(j+1)-1$}\Comment Access the column index for each nonzero entry
			\State $L(i,j) = L(i,j) \lor P(i,c(k))$ \Comment Entry is $\top$ if any of the entries accessed in $P$ are $\top$. 
			\EndFor
			\EndFor
			\EndFor
			\State \Return $L$
		\end{algorithmic} \caption{\label{alg:mat_mul} Sparse boolean matrix multiplication. Input: $c,r$, and $P$. Output: $L$ } 
	\end{algorithm}
	The advantage of using a sparse representation for $M$ is the reduction in memory requirements in proportion to the sparsity of $M$, and similarly the reduction in the number of operations required to perform the matrix multiplication.
	However, if the $\top$ entries are scattered throughout the matrix as illustrated in \eqref{eq:csr_demo}, the incremented values of $c(k)$ in line 3 of the multiplication algorithm will vary erratically leading to random memory accesses and poor cache utilization. 
	In light of assumption A3, the volume swept out by each trajectory is mapped to a clustered region within each row of the matrix remedying this problem.

	\begin{figure}
		\begin{Verbatim}[commandchars=\\\{\},frame=single,samepage=true]
\bcode{__global__} 
void matMult(uint32_t* c,
             uint32_t* r,
             bool* L_i, 
             bool* P_i,   
             uint32_t num_col)
  \{    
    uint32_t warp_id = blockIdx.x;
    uint32_t row_entries = c[warp_id+1] - c[warp_id]; 
    \bcode{__shared__} uint32_t col_id[THREADS_PER_BLOCK];
    \rcode{if}(warp_id >= num_col)
      return; 
    bool running = true;  
    L[warp_id] = false; 
    \rcode{for}(uint32_t i=0; 
        i<row_entries-THREADS_PER_BLOCK; 
        i+=THREADS_PER_BLOCK)
      \{
        \bcode{__syncthreads()};
        \rcode{if}(i+threadIdx.x < num_entries and running) 
          \{
            col_id[threadIdx.x] = c[r[warp_idx]+threadIdx.x+i];      
            \rcode{if}(P_i[col_id[threadIdx.x]]) 
              \{
                L_i[warp_id] = true;
                running = false;
              \}
            \bcode{__syncthreads()};
          \}
      \}
  \}
		\end{Verbatim}
		\caption{Kernel written in CUDA C for parallel evaluation of a sparse boolean matrix multiplied by a dense boolean vector. The matrix is stored in CSR format with the arrays c and r, while the dense boolean vector is stored in P\_i and the result is stored in L\_i. Each row of the sparse matrix is assigned a warp of GPU cores in an attempt to exploit the clustering of $\top$ (cf. Assumption A3) entries within each row and achieve coalesced memory access.}
	\end{figure}

\section{\label{sec:experiments}Application to Autonomous Freeway Driving}

These concepts are tested in the context of level 4 autonomous freeway driving where LTL specifications are used to represent safety requirements of freeway driving.

The dynamic model representing the mobility of the vehicle is the following:
\begin{equation}\label{eq:bicycle}
\begin{gathered}
\dot{p}_x = v\cos(\theta+\delta),\quad \dot{p}_y = v\sin(\theta+\delta), \quad \dot{\theta} = \frac{v}{l}\sin(\delta),
\\
\dot{v} = a,\quad \dot{\tau} = 1.
\end{gathered}
\end{equation}
This models the nonholonomic constraint of the single-track bicycle model with generalized coordinates $(p_x,p_y)$ located between the front wheels, and $\theta$ representing the heading of the vehicle. The steering angle $\delta$ and acceleration $a$ are treated as the control variables with longitudinal velocity $v$ integrating acceleration. 
Additionally, since autonomous driving requires accounting for dynamic objects, a state $\tau$ is used to keep track of time so that dynamic objects become static subsets of the resulting augmented state space.

\begin{figure}
	\centering
	\includegraphics[width=0.6\columnwidth]{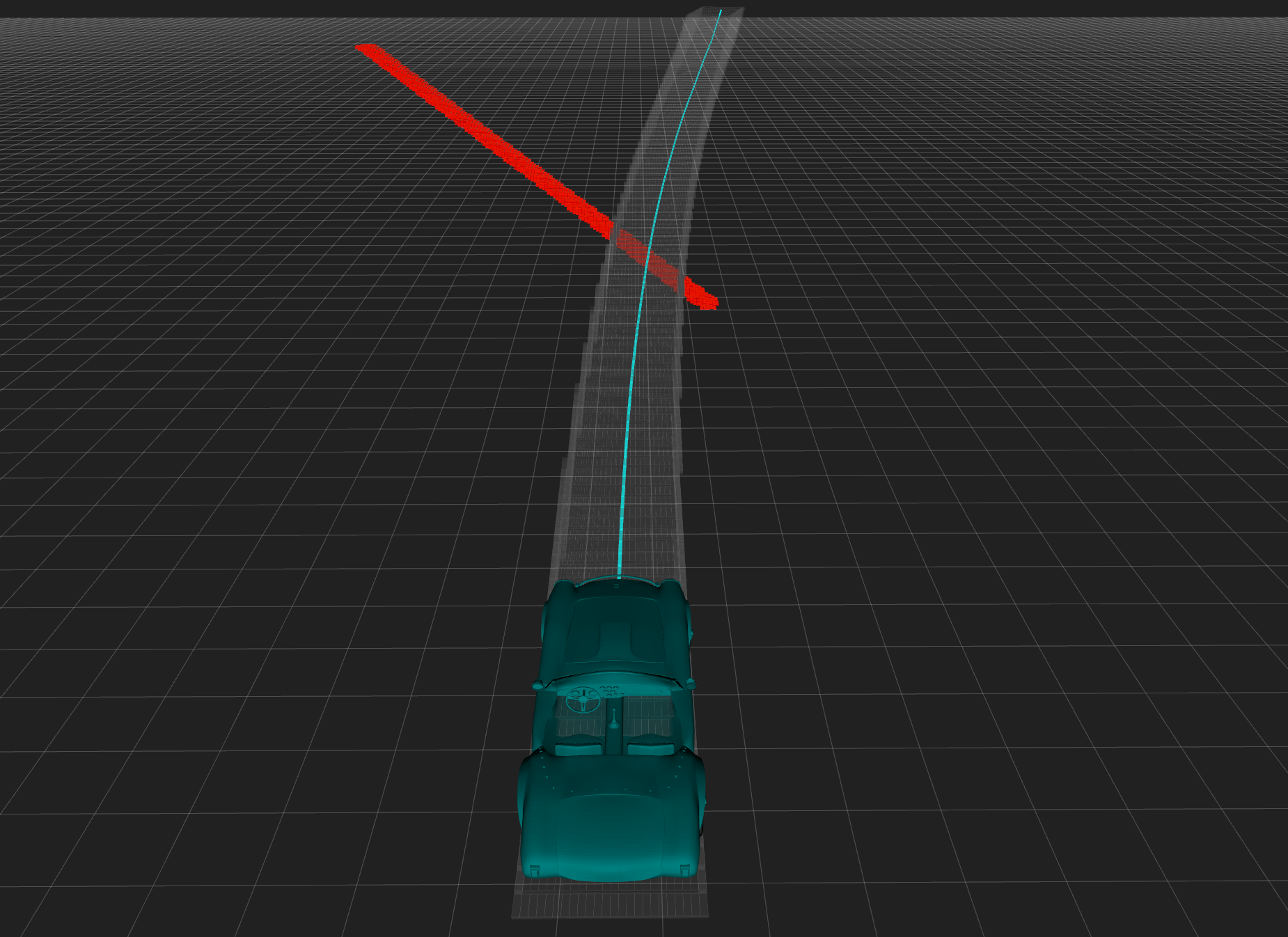}
	\caption{\label{fig:query} The white grid spans the $(p_x,p_y)$-plane, and the time coordinate $\tau$ spans the third dimension. 
		The blue curve shows the projection of a sample state trajectory in \eqref{eq:bicycle} into the workspace. 
		The opaque grey region is the volume swept out by the function $F$ in \eqref{eq:workspace_proj} associated with the volume covered by the footprint of the vehicle in the $(p_x,p_y)$-plane and time. 
		The 3D graphic of a vehicle is only an aid to understanding the swept volume depicted in the grey region. 
		The red volume represents a dynamic object moving across the path of the vehicle from right to left. 
		In the time-augmented workspace, an intersection-free maneuver is achieved with only a small change in the vehicle's speed and lateral position. 
		In contrast, without using time as a workspace dimension. The projection of the object into the $(p_x,p_y)$-plane would present a large obstacle making a sufficient lateral maneuver or longitudinal maneuver impossible given the vehicle's initial velocity.}
\end{figure}

The workspace $W$ consists of a rectangular subset of $\mathbb{R}^3$ with two coordinates associated with a position in the plane and the third coordinate is associated with time.
In reference to Section \ref{sec:acceleration}, the mapping $F$ from a state $(p_x,p_y,\theta,v,\tau)\in \mathbb{R}^5$ to a subset of the workspace is constructed by locating a top-down view of a bounding box or footprint of the vehicle determined by the coordinates $p_x$, and $p_y,$ and $\theta$.
This 2-dimensional rectangle is located at $\tau(t)=t$ along the third axis in the 3-dimensional workspace. 
To illustrate this, Figure \ref{fig:query} depicts the two subsets of the workspace that could be tested for intersection \eqref{eq:labeling_op}. 
In the time-augmented workspace the image of the state trajectory under $F$ does not encounter the dynamic object (proposition) which is traveling across the path of the vehicle. If a projection into the $(p_x,p_y)$-plane were used instead as the workspace,  the system would not be able to reason about the motion of the moving object.  

A discrete abstraction of the system in \eqref{eq:bicycle} is constructed using a probabilistic roadmap which uses a finite number of sampled states from the state space as states of the transition system with transitions to a number of each state's nearest neighbors computed by a steering function as required by the algorithm~\cite{karaman2011sampling}. 
From one planning query to the next, the vehicle's initial position may vary substantially. 
To remedy this issue, note that the dynamics in \eqref{eq:bicycle} are equivariant to translations along the $(p_x,p_y,\theta,t)$-subspace. 
That is, if the coordinate system is redefined as 
\begin{equation}
(p_x' ,p_y',\theta',v',t') \coloneqq (p_x + \Delta_x ,p_y + \Delta_y,\theta + \Delta_\theta, v, t+ \Delta_t),
\end{equation} 
for any value of $\Delta_*,$ then the transition system will remain dynamically feasible in the new coordinate system.

\subsection{Numerical Experiments}\label{sec:early_term}

\begin{figure}
\centering
\includegraphics[width=10cm]{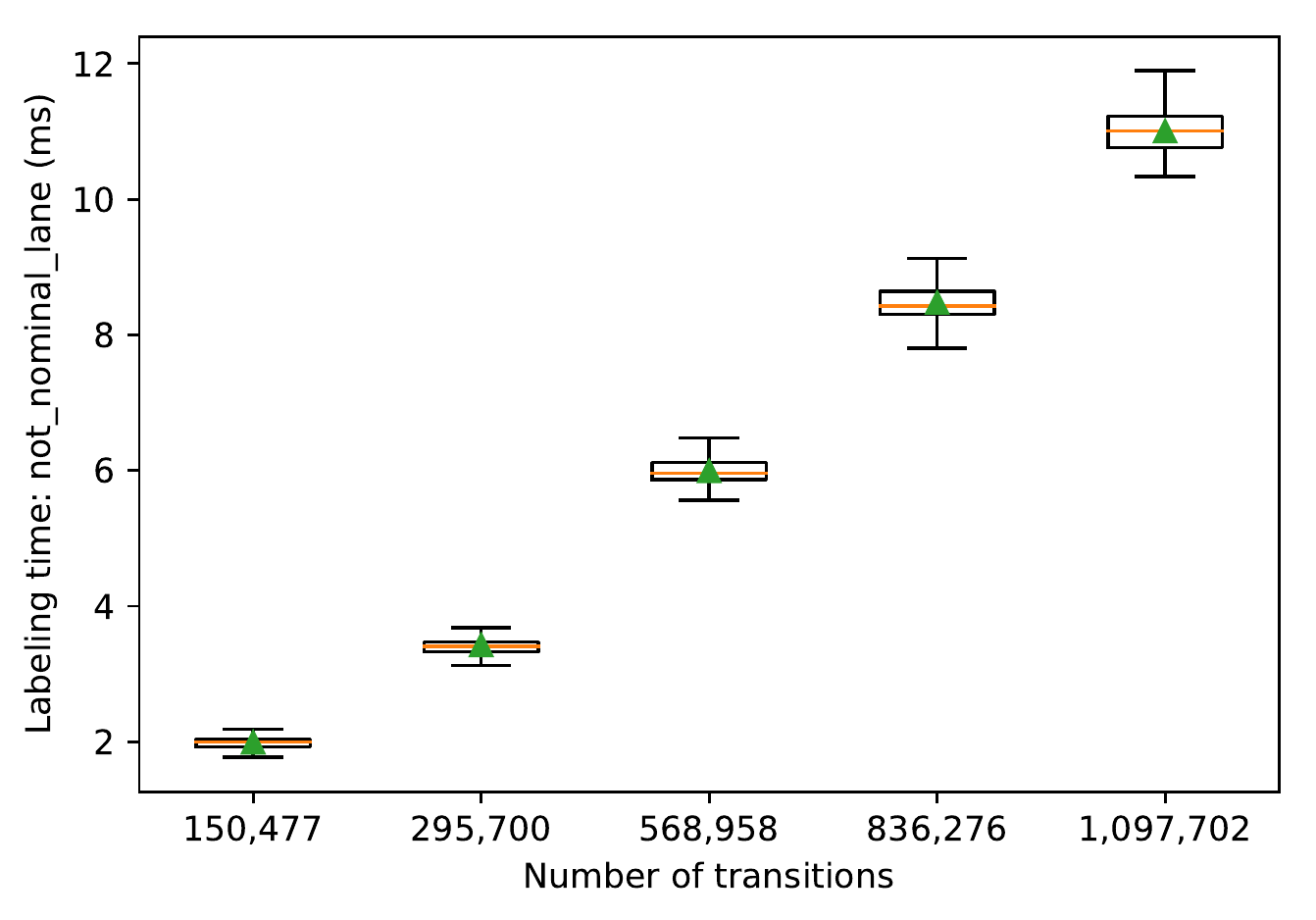}
\includegraphics[width=10cm]{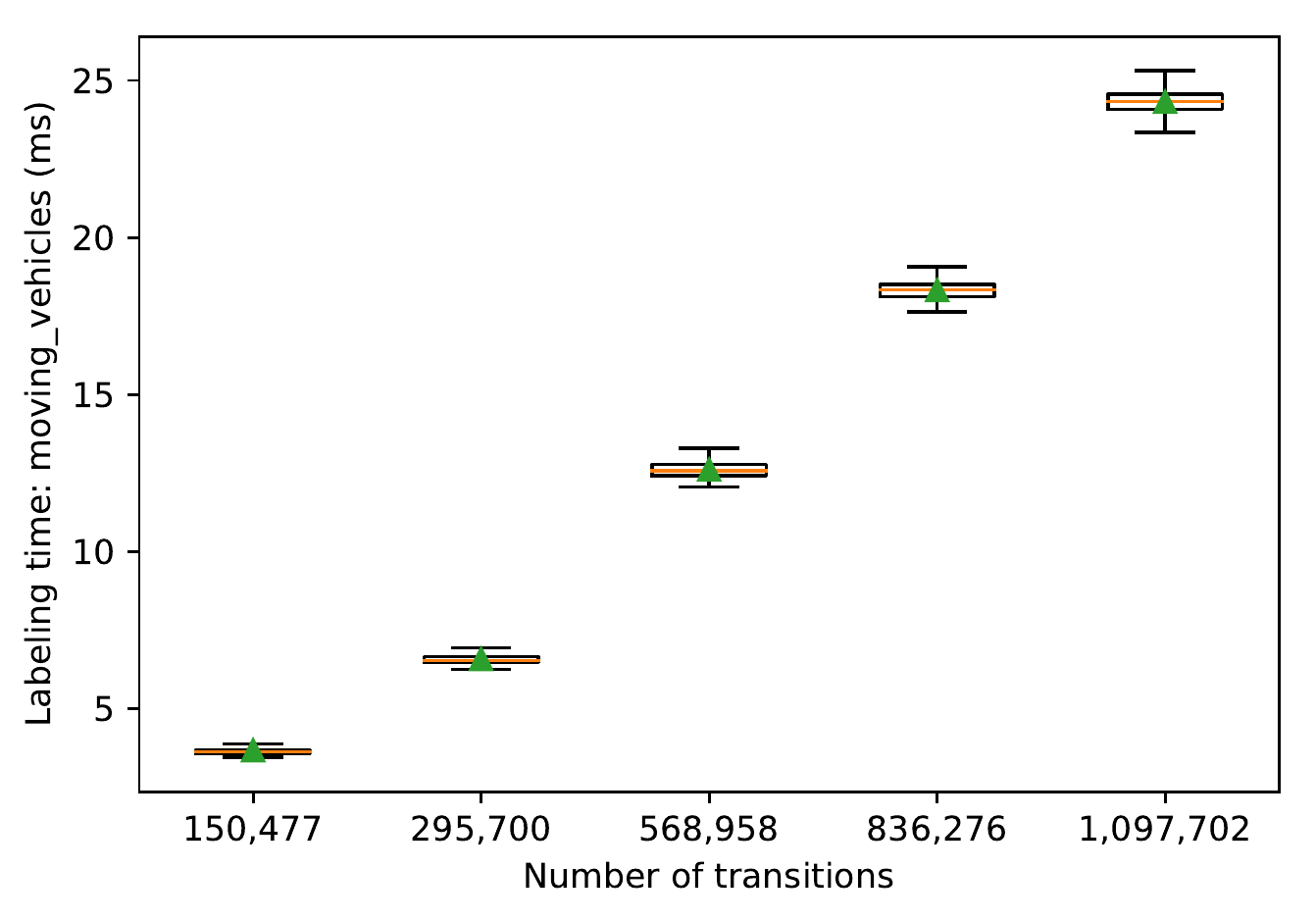}
\caption{\label{fig:chk-non-lane} Mean and variance of transition labeling time for the two propositions investigated. A linear dependence on the number of transitions is observed which is consistent with the number of binary operations required in \eqref{eq:bool_dot}. Labeling time is observed to be inversely related to proposition occupancy of the workspace. This is consistent, but not accurately predicted by equation \eqref{eq:prob_term} which justifies the need for realistic simulated scenarios.}
\end{figure}

The principal contribution of this paper is to present the performance of the proposed GPU-accelerated transition labeling procedure of Section \ref{sec:acceleration} on a transition system of interest to autonomous driving.
To generate realistic labeling function construction queries, a scenario was simulated where the vehicle travels on a closed circular loop with randomly generated agents driving along the route at various lateral positions and speeds. 
Two atomic propositions from the driving specification are tested.  
The first proposition $\mathtt{moving\_vehicle}$ represents the anticipated volume swept out by the other agents at each labeling function construction query.
The second proposition $\mathtt{not\_nominal\_lane}$ represents the region outside the nominal driving surface.
The subset of the workspace associated with $\mathtt{moving\_vehicle}$ represents dynamic objects occupying a small fraction of the workspace while $\mathtt{not\_nominal\_lane}$ represents a static object occupying a majority of the workspace's volume.
The workspace is partitioned into $2^{21}$ rectangular regions as described in Section \ref{sec:partition}, and an Nvidia GTX 1080 GPU was used for the experiments. The reported label construction times include the time to transfer data between the GPU over PCIe in addition to the computation running time.
Transition systems of various sizes are tested on 150 labeling function construction queries.
The labeling function construction results are summarized in Table \ref{tab:system-size}, and Figures \ref{fig:chk-non-lane}.

\paragraph{Importance of sampling test queries from realistic distributions:} 
Initial experiments were carried out on randomly generated subsets associated to motions and propositions. 
Each subset was generated by randomly selecting the occupancy of a voxel by sampling a bernoulli random variable with bias equal to the desired overall workspace occupancy.
This resulted in unrealistically fast computation time for the following reason: If the probability that a particular voxel is occupied by a motion or predicate is $p_{\mathrm{mot.}}$ and $p_{\mathrm{pred.}}$ respectively, then the probability that $n$ sequentially examined voxels are not simultaneously occupied by a motion and proposition is $(1-p_{\mathrm{pred.}} \cdot p_{\mathrm{mot.}})^n$ which tends to zero exponentially fast in $n$. 
If intersection is detected before all voxels have been examined, the remaining voxels have no effect on the result and the computation can be terminated early. 
 With random voxel selection, the number of voxels which need to be examined before early termination occurs follows a geometric distribution with an expected value,
 \begin{equation}\label{eq:prob_term}
	 \mathbb{E}[\mathrm{num.\,voxels\, compared}] = 1/(p_{\mathrm{pred.}} \cdot p_{\mathrm{mot.}}).
 \end{equation}
 
 %


%

\begin{table}
\caption{\label{tab:system-size} Transition system size and labeling function construction timescompute times}
\centering
\begin{tabular}{|c|c|c|c|c|c|}
\hline
Number of transtions & 154,776 & 295,700 & 568,958 & 836,276 & 1,097,702\\
\hline
Mean workspace occupancy (per transition) & 0.0223\% & 0.0223\% & 0.0223\% & 0.0223\% & 0.0223\% \\
\hline
labeling time ($\mathtt{moving\_vehicle}$) & 3.67ms & 6.60ms & 12.77ms & 18.46ms & 24.40ms \\
\hline
labeling time ($\mathtt{not\_nominal\_lane}$) & 2.01ms & 3.25ms & 5.95ms & 8.40ms & 11.02ms \\
\hline
Mean workspace occupancy ($\mathtt{not\_nominal\_lane}$) & 90.0\% & 89.9\% & 89.9\% & 89.9\% & 89.9\%
\\
\hline
Mean workspace occupancy ($\mathtt{moving\_vehicle}$) & 1.17\%  & 0.909\% & 0.928\%
 & 1.26\% & 1.07\% \\
\hline
GPU memory bandwidth utilization (Gb/s) & 177.61 & 188.55 & 187.68 & 190.96 & 189.62 \\
\hline
\end{tabular}
\end{table}


\subsection{Discussion}
In reference to Table \ref{tab:system-size}, the labeling function construction time scales linearly with the size of the transition system. 
This is consistent with the linear scaling of the number of binary operation required to compute the binary matrix multiplication in \eqref{eq:approx}. 
The $\mathtt{not\_nominal\_lane}$ proposition occupied roughly 90\% of the workspace while the $\mathtt{moving\_vehicle}$ proposition occupied only around 1\%.
One would expect that the proposition with higher occupancy would require less computation time as a result of the early termination phenomena discussed in Section \ref{sec:early_term}. 
This is consistent with the observation of roughly twice the time required for $\mathtt{moving\_vehicle}$ versus $\mathtt{not\_nominal\_lane}$.
However, the uniformly sampled random subset model \eqref{eq:prob_term} predicts a difference in amortized label construction time by a factor of 90:
\begin{equation}
\frac{1/(\num{9e-1} \cdot \num{2.23e-4})}{1/(\num{1e-2} \cdot \num{2.23e-4})} = 90.
\end{equation}
The difference between what is predicted and what is observed highlights the importance of sampling subsets for the performance study from realistic scenarios.

\section{Conclusion}
The proposed transition system labeling technique was demonstrated to label transition systems approximating the mobility of an autonomous vehicle at a rate of 4e7-8e7 transitions per proposition-second with variation due to the geometry and fraction of the workspace occupied by the subset associated to each proposition.
With a fairly standard PRM construction, a transition system with roughly 1e6 transitions is capable of demonstrating a wide range of maneuvers from low to freeway speeds.
If the driving specification is expressed with 10 atomic propositions, then the labeling task can be accomplished in 125ms-250ms. 
This is marginally suitable for autonomous driving where the latency of the entire system must be well under one second.
The implementation is not highly optimized for the GPU architecture however and with some effort the performance could likely be improved by considering more effective use of shared memory and caches.
Alternatively, an application specific integrated circuit could be constructed to perform the labeling operation with a greater degree of parallelism.
%

\bibliographystyle{splncs}
\bibliography{references}

\end{document}